**One-dimensional convolutional neural network model for breast cancer subtypes classification and biochemical content evaluation using micro-FTIR hyperspectral images.**

Matheus del-Valle[a], Emerson Soares Bernardes[b], Denise Maria Zezell[c,*]

[a] Lasers and Applications Center, Energy and Nuclear Research Institute, University of São Paulo, São Paulo, Brazil

[b] Radiopharmacy Center, Energy and Nuclear Research Institute, University of São Paulo, São Paulo, Brazil

* Corresponding author.

E-mail addresses: matheusdv@gmail.com (M. del-Valle), emerson.bernardes@gmail.com (E.S. Bernardes), zezell@usp.br (D.M. Zezell).

**Abstract**

Breast cancer treatment still remains a challenge, where molecular subtypes classification plays a crucial role in selecting appropriate and specific therapy. The four subtypes are Luminal A (LA), Luminal B (LB), HER2 subtype, and Triple-Negative Breast Cancer (TNBC). Immunohistochemistry is the gold-standard evaluation, although interobserver variations are reported and molecular signatures identification is time-consuming. Fourier transform infrared micro-spectroscopy with machine learning approaches have been used to evaluate cancer samples, presenting biochemical-related explainability. However, this explainability is harder when using deep learning. This study created a 1D deep learning tool for breast cancer subtype evaluation and biochemical contribution. Sixty hyperspectral images were acquired from a human breast cancer microarray. K-Means clustering was applied to select tissue and paraffin spectra. CaReNet-V1, a novel 1D convolutional neural network, was developed to classify breast cancer (CA) and adjacent tissue (AT), and molecular subtypes. A 1D adaptation of Grad-CAM was applied to assess the biochemical impact to the classifications. CaReNet-V1 effectively classified CA and AT (test accuracy of 0.89), as well as HER2 and TNBC subtypes (0.83 and 0.86), with greater difficulty for LA and LB (0.74 and 0.68). The model enabled the evaluation of the most contributing wavenumbers to the predictions, providing a direct relationship with the biochemical content. Therefore, CaReNet-V1 and hyperspectral images is a potential approach for breast cancer biopsies assessment, providing additional information to the pathology report. Biochemical content impact feature may be used for other studies, such as treatment efficacy evaluation and development new diagnostics and therapeutic methods.

**Keywords:** deep learning; 1D CNN; micro-FTIR; hyperspectral imaging; breast cancer; molecular subtype; biochemical content

**1. Introduction**

Breast cancer is the most common cancer worldwide, accounting for 11.7% of new cancer cases and 6.9% of cancer deaths in 2020 alone [1]. Despite advancements in classification methods, such as stage, grade, and molecular subtypes, breast cancer treatment still remains a challenge. Molecular subtypes classification, based on the expression levels of various biomarkers, plays a crucial role in selecting appropriate and specific therapy [2]. The four subtypes are luminal A, luminal B, human epidermal growth factor receptor 2 subtype (HER2), and triple-negative breast cancer (TNBC) [3,4].

Histology and immunohistochemistry are the gold-standard to classify breast cancer subtypes, although variations in antibodies, detection systems, and protocols can affect the accuracy of these techniques [5]. Interobserver variations are also reported, where the identification of molecular signatures is time-consuming, besides expensive [6]. Researchers have explored alternative approaches, including the Fourier Transform Infrared (FTIR) spectroscopy, as it may supply additional content information and improve assessment quality [7,8] due to the biochemical information brought by this technique, such as lipids, proteins, nucleic acids and carbohydrates contents [9].

Machine learning approaches have been applied to micro-FTIR hyperspectral imagens to assist in cancer evaluation, as it is possible to predict important parameters along a biochemical related explainability [10–12]. However, such explainability is harder to be achieved for one of the main machine learning tools, the deep learning, where several efforts have been performed to decrease its black box characteristic [13,14].

While several studies apply deep learning for vibrational spectroscopy [15], there is no study so far classifying breast cancer molecular subtypes of breast cancer biopsies and analyzing the biochemical content impact for such classification, specially using deep learning approaches. In this way, this study evaluated a one-dimensional deep learning approach to identify the breast cancer presence and to classify the molecular subtype, while pointing out the biochemical contribution that influenced the classification. This automated information may provide extra information for the pathology report and for a better understanding of the breast cancer.

## 2. Material and Methods

### 2.1. Dataset

The BR804b (Biomax, Inc, USA) breast cancer microarray was ordered with formalin fixation and paraffin embedding (FFPE) histological sections of 8 µm, fixed in calcium fluoride ($CaF_2$) slides (Crystran, UK). A total of 60 cores of 1.5 mm were imaged, one Cancer and one Adjacent Tissue (AT) core for each of the 30 unique patients. Cores were already classified by Biomax regarding their type, and receptors and Ki67 expression levels. Molecular subtypes were classified based on St. Gallen International Expert Consensus guidelines [3,4], thus resulting in the distribution (where N denotes the samples number): Type – Cancer (CA; N=30) and Adjacent

Tissue (AT; N=30); Subtype – Luminal A (LA; N=8), Luminal B (LB; N=8), HER2 (N=7) and Triple-negative Breast Cancer (TNBC; N=7).

Image acquisition was performed using a Cary Series 600 system (Agilent Technologies, USA), composed by a Cary 660 FTIR spectrometer and a Cary 620 FTIR microscope. It was used the focal plane array (FPA) detector to acquire 320x320 pixels hyperspectral images with 5.5 µm spatial resolution for each core, resulting in 6,144,000 raw spectra. Tissue and borderline paraffin were gathered. The system was set to operate between 3950 and 900 cm$^{-1}$, with 4 cm$^{-1}$ spectral resolution in transmission mode. Background images were acquired with 256 co-added scans, while 64 scans were set to the sample images acquisition.

A single 320x320 image of environment spectra was collected using a clean slide to obtain water vapor (H2O) variation. The air purge of the FTIR microscope acrylic box was turned off and the box left open when this acquisition started to increase the H2O variation.

*2.2. Data preprocessing*

Preprocessing steps were applied individually to each image. Tissue and paraffin regions were selected using a two-step K-means clustering, while pure slide spectra were excluded. Firstly, for the tissue identification, raw spectra were truncated at the Amide I and II region (1700 to 1500 cm$^{-1}$) and modeled with k=2. Secondly, for the paraffin identification, raw spectra were truncated at the highest paraffin intensity band (1480 to 1450 cm$^{-1}$) and also modeled with k=2, but spectra previously clustered as tissue were set to zero intensity. An area integration check was performed to guarantee tissue and paraffin as cluster 1 in each K-means.

A biofingerprint truncation was applied from 1800 to 900 cm$^{-1}$, resulting in 467 points. Outlier removal was applied using the Hotelling's $T^2$ vs Q residuals method, with 10 Principal Components (PC) and a 95% confidence interval fixed threshold removal. Data were smoothed using Savitzky-Golay (SG) filtering with window size of 11 and polynomial order of 2. These steps were also applied to the H2O spectra.

Extended Multiplicative Signal Correction (EMSC) [16] was applied coupled with paraffin removal [17], from 1500 to 1350 cm$^{-1}$, and water vapor (H2O) removal, from 1800 to 1300 cm$^{-1}$. The EMSC model was built by a polynomial baseline of order 4; PCA from paraffin and H2O using the number of PCs that corresponded to 99% of explained variance; and global mean spectra from tissue, paraffin, and H2O as references. The model was solved by least squares estimation. Spectra were normalized by min-max method and another outlier removal was performed using Hotelling's T$^2$ vs Q residuals method. Preprocessed spectra were saved in a HDF5 file. Spectra were labeled as binary encoding for the type classification, while the subtype was handled by a one-hot encoding.

*2.3. Deep learning*

A novel 1D convolutional neural network (CNN) called CaReNet-V1 was developed based on VGG [18], ResNet [19,20] and reported 1D models for spectroscopy analysis [21–24]. **Fig. 1** presents the CaReNet-V1 architecture.

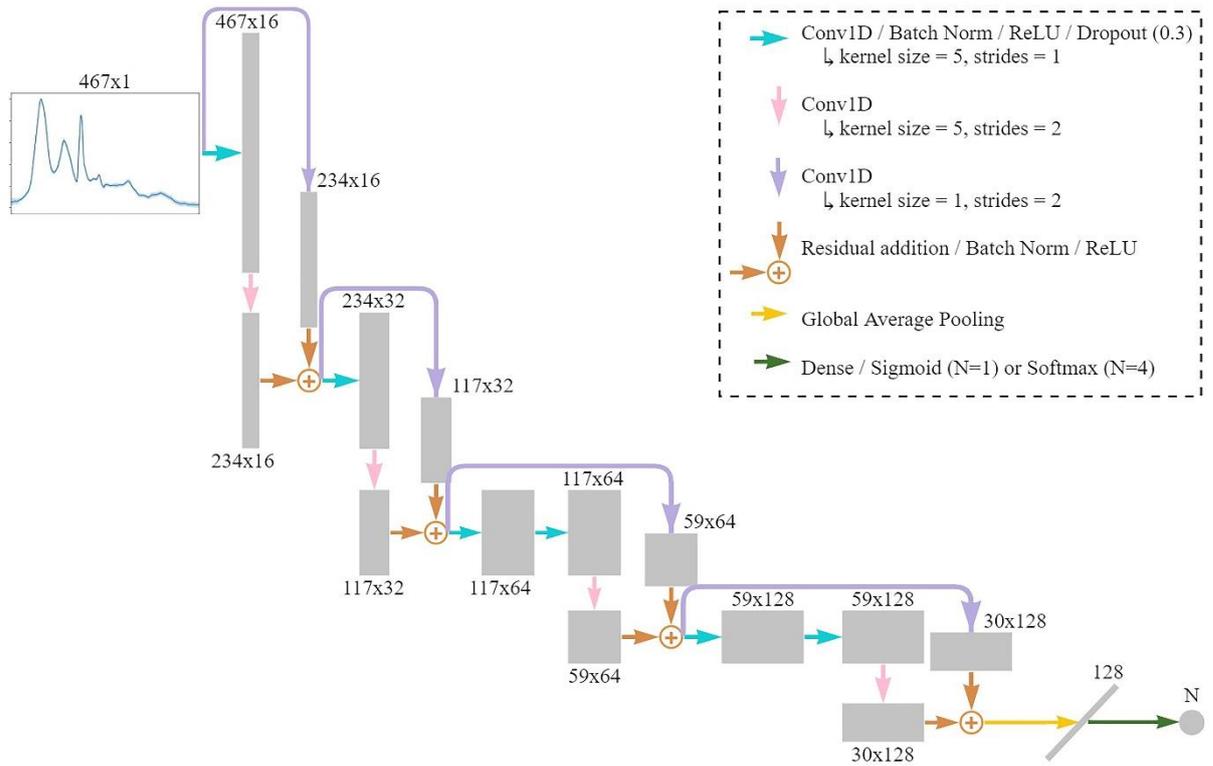

**Fig. 1**. CaReNet-V1 architecture.

All convolutional layers were created using HeNormal kernel initialization [25] and zero padding. The type classification model was created with a single final neuron, sigmoid activation and binary cross-entropy loss, while subtype one was built using four final neurons with softmax activation and categorical cross-entropy loss. Models were trained using Adam optimizer [26], with learning rate of 1e-3, beta 1 of 0.9 and beta 2 of 0.999. A reduce learning rate on plateau callback [27] was employed to monitor the testing loss, with patience of 4, reduce factor of 0.5 and minimum learning rate of 1e-4.

A total of 4 patients were held-out for the test set, one for each subtype class and two for each type class. The 26 remaining patients were addressed for a balanced stratified 4-fold cross-validation, resulting in 21 patients for the train set and 5 for the development (dev) set of each fold, except for the last one, containing 20/6 (train/dev). The balance was executed by randomly undersampling the training spectra until reaching the same quantity per class. A batch size of 250 spectra was defined, along 50 training epochs. The training spectra order was randomly shuffled for each epoch using a data generator.

Performance was evaluated by assessing each spectrum prediction and the sample prediction, where the most predicted class among all sample spectra was determined as the final classification. A 0.5 threshold was applied for type classification and the maximum argument was chosen as the subtype prediction. The evaluation was accomplished by accuracy, specificity and sensitivity metrics, and by a 1D adaptation of Gradient-weighted Class Activation Mapping (Grad-CAM) [28]. Final Grad-CAM was calculated using the best fold model from dev set and averaging the Grad-CAM of each sample, grouping by classes. A min-max normalization was applied to generate the Grad-CAM heatmap. All the study was performed by in house algorithms in Python, mainly Tensorflow and Keras libraries, and using a GeForce GTX 1080 GPU with 8 GB of memory.

## 3. Results and Discussion

### 3.1. Data preprocessing

**Fig. 2** shows a representative image of the clustering process. The Amide I peak image in **Fig. 2 (b)** is only for visual comparison, where it is possible to evidence its relation to the white light image in **Fig. 2 (a)**. Despite that, a single peak may not a good approach to cluster the spectra, since relying on more information is more reliable.

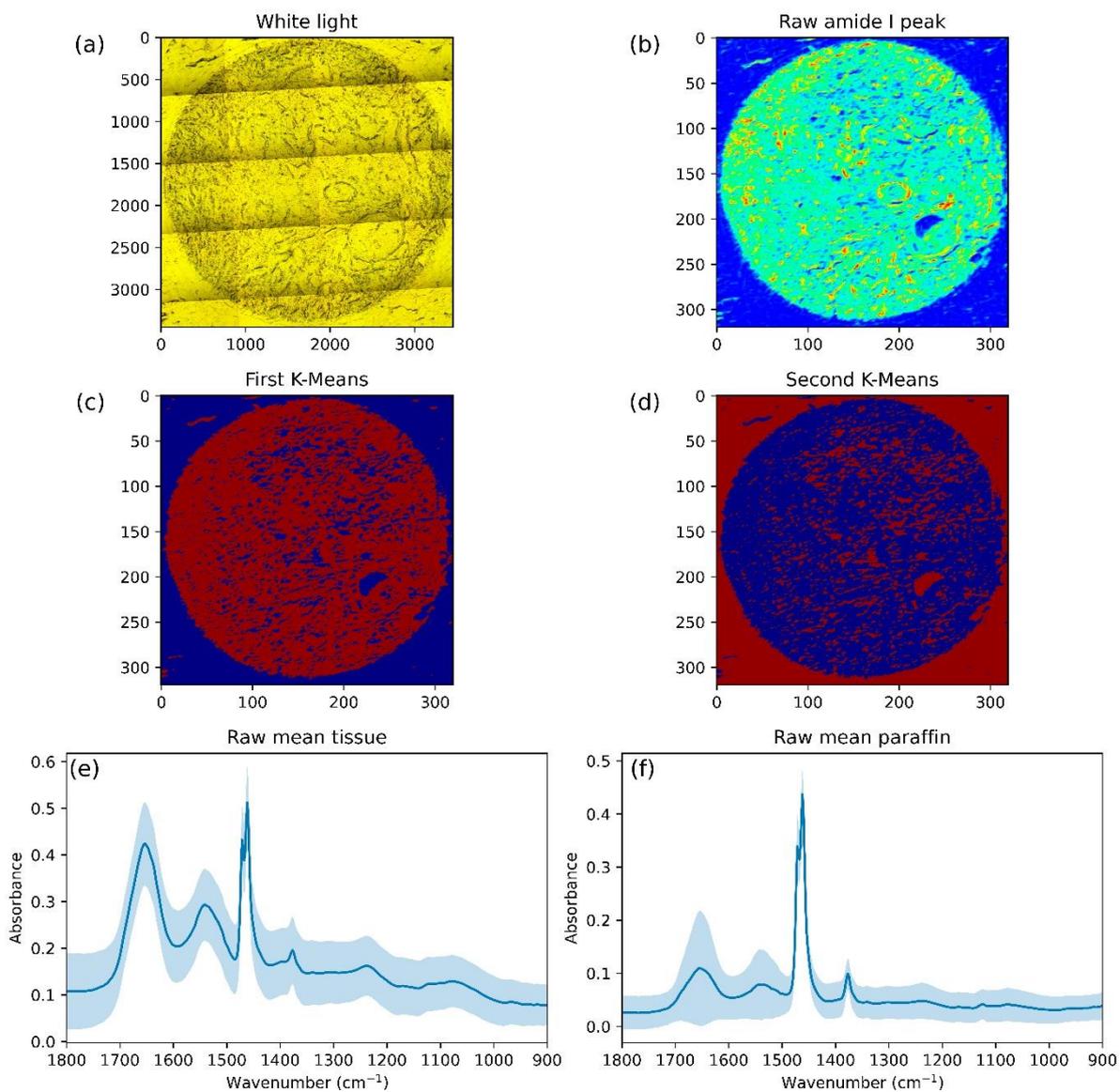

**Fig. 2.** Representative figure of the clustering process. (a) White light image acquired by Cary microscope in transmission mode; (b) Amide I intensity peak plot from the raw spectra; (c) First K-Means clustering, evidencing tissue regions in red; (d) Second K-Means clustering, evidencing paraffin regions in red; (e) Raw mean spectrum (solid line) and standard deviation (light shadow) from tissue regions in (c); (f) Raw mean spectrum (solid line) and standard deviation (light shadow) from paraffin regions in (d). Spatial scale of images (a) to (d) in pixels.

Amide bands are indicators of biological tissue presence, as other regions such as paraffin and pure slide does not present these bands [29]. Therefore, the first K-Means clustering using Amide bands was able to select tissue related spectra as Cluster 1 (red), leaving paraffin and any pure slide spectra as Cluster 0 (blue), as showed in Figure **Fig. 2 (c)**. The raw spectrum in Figure **Fig. 2 (e)**, identified as Cluster 1 by the first K-Means, exhibits the pattern of an usual biological tissue spectrum, being possible to evidence the characteristic Amide I and II bands (1700 to 1500 $cm^{-1}$), as well as the paraffin bands, since the tissue is embedded in paraffin, with peaks in ~1462 $cm^{-1}$ and ~1373 $cm^{-1}$ [9].

If it was guaranteed no pure slide spectra in the acquisition, only the first K-Means would be enough, however, some regions may present the absence of tissue and paraffin. In this way, a second K-Means is necessary, being able to identify paraffin spectra as Cluster 1 (red) as in **Fig. 2 (d)**. As tissue spectra identified in the first clustering were set to zero, tissue and pure slide spectra should be similar with low or absent signal at the highest paraffin intensity band (1480 to 1450 $cm^{-1}$), being both selected as Cluster 2 (blue).

The raw spectrum in Figure **Fig. 2 (f)**, identified as Cluster 1 by the second K-Means, shows the selected paraffin. It is possible to evidence an intensity variation in the Amide I and II bands, which are not related to pure paraffin. This may be due to thinner tissue regions from the histological sectioning, hence not presenting enough Amide intensities to be clustered as tissue, but as paraffin.

Small contribution of tissue regions in the paraffin spectra, does not affect non-paraffin regions of the final tissue during the EMSC modeling, since the paraffin model masks the region from 1500 to 1350 $cm^{-1}$. Although this contribution my affect the variance of the masked paraffin region, the PCA model in the EMSC should take more into account the paraffin variance. Analogously to paraffin, H2O was masked from 1300 $cm^{-1}$ onward and, therefore, only this region accounted for the EMSC model. In addition, outlier removal techniques assist to overcome these issues.

**Fig. 3** presents a final spectrum, using the sample from **Fig. 2**, with the biochemical regions identification [30,31]. The smaller standard deviation, covered by the spectrum midline in some regions, is a result of the preprocessing steps, making the tissue spectra comparable.

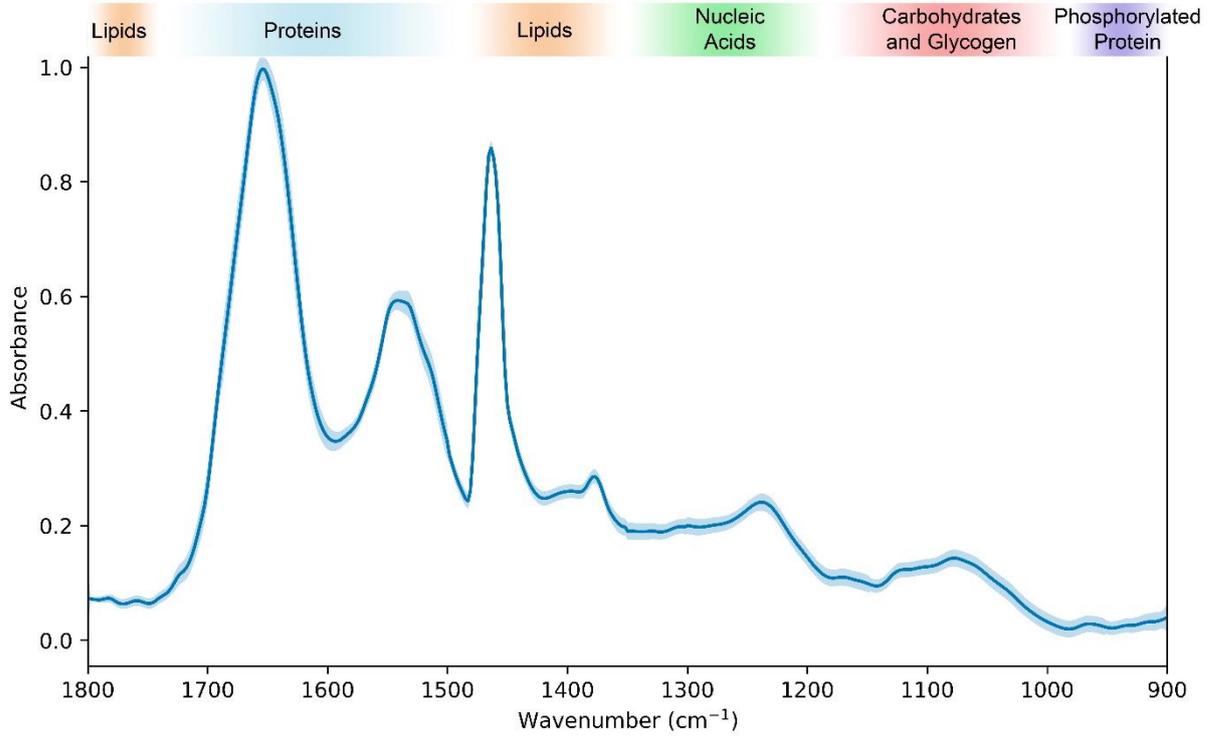

**Fig. 3**. Representative final tissue spectrum with the corresponding biochemical regions. Mean spectrum (solid line) and standard deviation (light shadow).

*3.2. Deep learning*

**Table 1** shows the dev and test sets performances for each class. The type classification presented higher values than the subtype, besides closer values of dev and test, while subtype demonstrated lower test values in comparison with dev values. This indicates the type classification as the easiest prediction to be learned, as it was expected, since cancer and AT are the most different samples analyzed.

**Table 1**

CaReNet-V1 dev and test performance. Results grouped by classes of each model (type and subtype). Mean values ± standard deviation.

| Set | Label | Class | Accuracy | Specificity | Sensitivity |
|---|---|---|---|---|---|
| Dev | Type | CA | 0.95 ± 0.02 | 0.92 ± 0.03 | 0.97 ± 0.02 |
| | Subtype | LA | 0.89 ± 0.04 | 0.83 ± 0.05 | 0.82 ± 0.08 |
| | | LB | 0.87 ± 0.05 | 0.83 ± 0.09 | 0.79 ± 0.05 |
| | | HER2 | 0.93 ± 0.03 | 0.91 ± 0.02 | 0.92 ± 0.04 |
| | | TNBC | 0.92 ± 0.02 | 0.87 ± 0.05 | 0.89 ± 0.07 |
| Test | Type | CA | 0.89 ± 0.03 | 0.89 ± 0.02 | 0.93 ± 0.03 |
| | Subtype | LA | 0.74 ± 0.08 | 0.45 ± 0.09 | 0.62 ± 0.06 |
| | | LB | 0.68 ± 0.09 | 0.61 ± 0.10 | 0.42 ± 0.11 |
| | | HER2 | 0.83 ± 0.05 | 0.89 ± 0.02 | 0.78 ± 0.06 |
| | | TNBC | 0.86 ± 0.03 | 0.85 ± 0.04 | 0.92 ± 0.03 |

LA and LB metrics were the lowest, however, dev results exhibit that the model was able to learn how to extract features of the spectra to predict the classes. Nevertheless, test results show specificity and sensitivity near 0.5, indicating the model had more difficulties with this class. Even though several spectra predictions were wrong, the final test patient could be right, as it is the most predicted class, where **Table 2** demonstrates this performance.

**Table 2**

CaReNet-V1 performance for each test patient and each of the four models from the folds in relation to the GT (Ground Truth). Light blue indicates correct predictions, while light red are wrong predictions.

| Label | GT Class | Predicted class – Model fold: | | | |
|---|---|---|---|---|---|
| | | 1 | 2 | 3 | 4 |
| Type | AT | AT | AT | AT | AT |
| | AT | AT | CA | AT | AT |
| | CA | CA | CA | CA | CA |
| | CA | CA | CA | CA | CA |
| Subtype | LA | LB | LA | LA | LA |
| | LB | LA | LA | LB | LB |
| | HER2 | HER2 | HER2 | HER2 | LA |
| | TNBC | TNBC | TNBC | TNBC | TNBC |

The model efficiently classified the type, as verified in the single spectra performance, with only one AT being classified as CA. This negative impact can be reduced when considering that it is better to have a false positive than a false negative for a cancer evaluation.

The model struggled to differentiate LA and LB samples. Luminal subtypes are the most similar, as they can even present the same receptors expression, being different only regarding the Ki67 level [4]. HER2 and TNBC were correctly classified in all cases, except by one HER2 as LA.

The concern of distinguishing LA and LB is lower than finding HER2 and TNBC cases, once they are more aggressive manifestations of the breast cancer than the luminal ones [32]. Perfect TNBC prediction is especially important as this subtype exhibits the worst prognosis due to lack of drug targets and high risk of brain metastasis [33]. Therefore, these findings can be considered an important step towards an automated laboratory screening technique, and may help to prioritize patients' analysis by the health professionals.

Grad-CAM results are depicted in **Fig. 4** and **Fig. 5**. This is a localization approach to identify important regions of the image that influenced the classification decision [28]. Once CaReNet-V1 is a 1D model, the localization is related to the wavenumber instead of the spatial information. Thus, a 1D Grad-CAM can be applied as a feature importance tool.

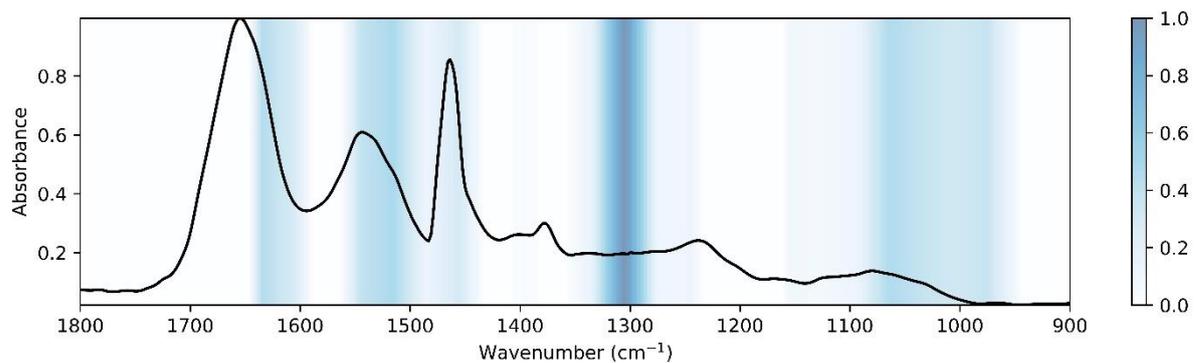

**Fig. 4.** Gradient-weighted Class Activation Mapping (Grad-CAM) of the Type classification model. Darker blue areas identify wavenumber regions that contributed the most for the Cancer (CA) activation.

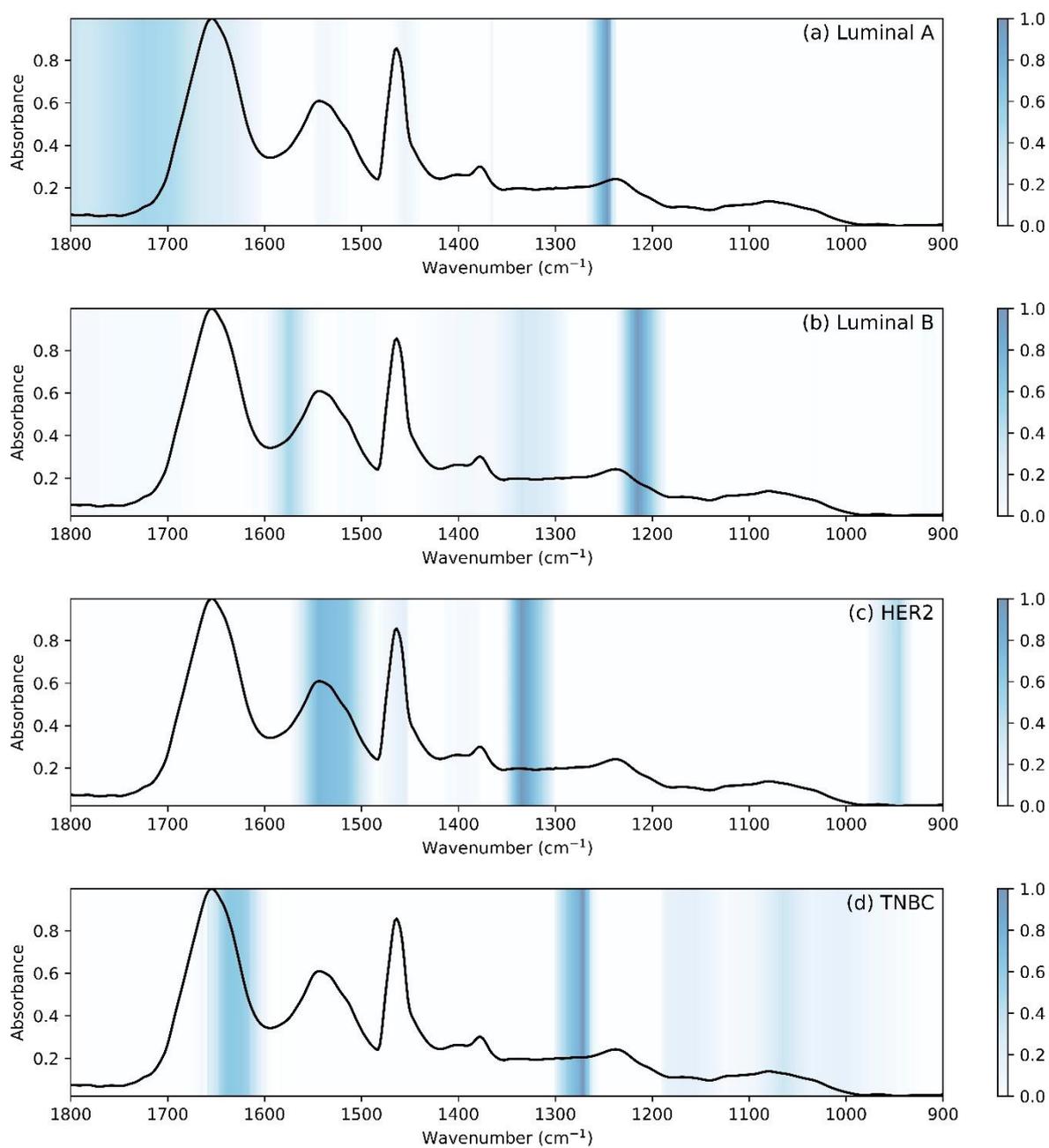

**Fig. 5** Gradient-weighted Class Activation Mapping (Grad-CAM) of the Subtype classification model. Darker blue areas identify wavenumber regions that contributed the most for the activation of (a) Luminal A (LA), (b) Luminal b (LB), (c) HER2, and (d) Triple-Negative Breast Cancer (TNBC).

Type classification was performed by one output (one neuron with sigmoid activation), thus only one Grad-CAM can show the activation or not of the CA class. On the other hand, subtype with four output probabilities leads to four Grad-CAM. These results can be related to the molecular footprints of vibrational spectroscopy [31,34], as per the following, to understand the composition impacts to the model.

Cancer type activation was mainly related to four regions. The first one in 1640-1600 cm$^{-1}$ is mainly assigned to adenine vibrations in DNA and one part of Amide I. Second region of 1550-1510 cm$^{-1}$ is strongly related to Amide II band, where there is C–N stretching and C–N–H bending vibrations weakly coupled to the C=O stretching mode, besides the nucleobases C=N. Studies have reported amide I and II to differentiate cancerous and healthy tissue [35]. The third, in 1320-1280 cm$^{-1}$ band, there is the leading contribution of Amide III, essentially related to C–N stretching, N–H in plane bending and CH$_2$ wagging vibrations; and collagen associated vibrations. The most intense activation contribution in 1310-1300 cm$^{-1}$ is totally related to Amide III. At last, the 1080-980 cm$^{-1}$ region is pertinent to glycogen (1050-1020 cm$^{-1}$) and symmetric PO$_2^-$ stretching in RNA and DNA (1100-1040). Glucose expressions have been linked to cancer cells during the neoplastic process, while DNA and RNA oscillation is directly associated with cancer diagnosis [36].

Regions that contributed the most for LA classification (**Fig. 5 (a)**) are around 1750-1680 and 1260-1240 cm$^{-1}$. The first one is mostly related to C=O stretch from bases of nucleic acids in 1717-1681 cm$^{-1}$, and from lipids and fatty acids in 1750-1725 cm$^{-1}$. It is stated that tumor progression and cancer cell survival favored by fatty acids overproduction [37]. In the second region, there is the contribution of Amide III, phosphate, and collagen I and IV. Amide III region is observed in 1340-1240 cm$^{-1}$ due to C–N stretching of proteins, indicating mainly α-helix conformation. In 1245-1240 cm$^{-1}$ there are several PO$_2^-$ asymmetric stretching originated from the phosphodiester groups of nucleic acids, which suggest nucleic acids increase in malignant tissues

[31]. Presence of $PO_2^-$ stretching vibrations may be associated to DNA damage caused by reactive oxygen species [36].

LB classification contribution (**Fig. 5 (b)**) from 1590-1570 cm$^{-1}$ relates to C=N adenine and to phenyl ring C−C stretch. The accelerated metabolism of DNA/RNA leads to oscillatory deformations of adenine [36]. The 1225-1200 cm$^{-1}$ region is associated with $PO_2^-$: asymmetric phosphate I vibrations at 1217-1207 cm$^{-1}$; asymmetric phosphate vibrations of nucleic acids when highly hydrogen-bonded; and phosphate II asymmetric vibration in B-form DNA.

Three regions appear with high impact for HER 2 classification (**Fig. 5 (c)**): 1550-1510, 1350-1300, and 980-950 cm$^{-1}$. The first region is all covered by the amide II band, with N−H bending vibration coupled to C−N stretching. Guanine C=N is also present in 1534-1526 cm$^{-1}$. In 1317-1307 cm$^{-1}$ there are amide III band components of proteins, and in 1340-1317 cm$^{-1}$ collagen related assignments. Last region is mainly attributed to deoxyribose C−O and symmetric stretching mode of dianionic phosphate monoesters in phosphorylated proteins and nucleic acids. Increased intensities in this region are correlated to cells in malignant tissue [34].

TNBC classification (**Fig. 5 (d)**) showed the influence in 1660-1610 cm$^{-1}$, which covers a large area of the amide I band, besides adenine vibration in DNA in 1609-1601 cm$^{-1}$. In 1300-1270 cm$^{-1}$, there are mainly amide III and collagen vibrational modes, and also a $CH_2$ wagging vibration of phospholipids acyl chains. Phospholipids expression is used to distinguish subtypes membrane remodeling, where TNBC and HER2 demonstrates the greater difference and potentially reflects their greater ability to grow [37]. The presented vibrational modes of the collagen are distinctly stronger for breast carcinoma [31]. A slight contribution in 1070-1050 cm$^{-1}$ is mainly associated to phosphate and oligosaccharides, such as P−O−C antisymmetric stretching and C−OH stretching.

In this way, the 1D deep learning prediction coupled with a Grad-CAM analysis of micro-FTIR images can assist to understand the breast cancer composition that distinguish the cancer and subtypes in a label-free approach. Although wavenumber shifts may occur according to the sample, the band (region) analysis offers more reliable evaluation. This analysis may be employed not only for diagnosis purposes, but also for treatment efficacy assessment and development of new therapeutic methods.

Models created with a 1D approach have the advantage of consuming less memory, hence being easier to be trained. CaReNet-V1 architecture, using 467x1 input shape, resulted in 277,236 parameters. In contrast, albeit one spectrum prediction is fast, as there are several spectra in one mosaic, it takes longer to predict one whole patient. One-dimensional adaptations of traditional CNN, such as VGG, and models based on the 1D spectroscopy analysis reported in [21–24] were tested, yet the learning process was only possible when using residuals approaches as in ResNets, where CaReNet-V1 model presented best results.

Using single spectra increases the dataset size, where one mosaic of 320x320 turns into 102.400 single spectra, facilitating the training with even a couple of mosaics. Even so, training with several patients is recommended to enable a generalist model. In addition, the single spectra approach raises the issue that not all spectra may be representative of the class, as a breast cancer biopsy usually presents a very heterogeneous tissue. This can directly decrease the 1D model performance, since all spectra is considered as being from the same class, while still correctly predicting the final patient classification due to the incorporate voting system.

The second derivative was tested as input instead of the regular absorbance, and also a double channel input with absorbance and its second derivative, but in all cases using only absorbance resulted in better outcomes. Hence, second derivative results were omitted from this study for simplicity. Analogously, a cosine decay restarts schedule [38] was tested, but no improvement was observed. While it enables longer training, such as 500 epochs, to increase the probability of the optimizer to reach the global minimum loss, the reduce learning rate on plateau callback granted similar performance with only 50 epochs.

Next studies should consider adding more breast cancer biopsies. Increasing the dataset size may help not only by giving more training and dev examples, but mainly by augmenting the test set quantity, hence aiding to achieve a better real world performance evaluation of the breast cancer subtypes. In addition, individual receptor expressions should be assessed along biopsies augmentation, once some tests with the current dataset have demonstrated poor performance of CaReNet-V1 for these predictions.

## 4. Conclusion

The clustering method using two K-Means was able to identify tissue and paraffin spectra, enabling an automated data organization and preprocessing. The individual spectra processing of the 1D approach augmented dataset size, where a single 320x320 pixels mosaic resulted in 102.400 spectra.

CaReNet-V1 efficiently classified breast cancer tissue (CA) against adjacent tissue (AT), with only one test patient false positive. Subtypes were correctly classified, except for four Luminal A/Luminal B wrong predictions. The 1D model coupled with a Gradient-weighted Class Activation Mapping (Grad-CAM) enabled a detailed evaluation of the feature importance, directly correlated to the biochemical composition of the samples, which may assist to better understand the subtypes composition, diagnosis and therapeutic approaches.

A larger dataset may assist to evaluate other parameters, as biomarkers levels and other histopathology evaluations, while maintaining the biochemical content evaluation. The findings in this study indicates the novel 1D deep learning and micro-FTIR imaging a potential approach for breast cancer biopsies assessment, providing additional information to the pathology report. In addition, biochemical content impact feature may be used for other studies, such as treatment efficacy evaluation and development new diagnostics and therapeutic methods.

**Acknowledgements**

The authors would like to thank the Health Innovation Techcenter of Hospital Israelita Albert Einstein for sharing their GPU workstation.

**Conflict of interest**

The authors do not have any conflicts of interest to disclose.

**Funding**

This work was supported by FAPESP (CEPID 05/51689-2, 17/50332-0), CAPES (Finance Code 001, PROCAD 88881.068505/2014-01) and CNPq (INCT-465763/2014-6, PQ-309902/2017-7, 142229/2019-9).